\documentclass[final]{cvpr}

\usepackage{times}
\usepackage{epsfig}
\usepackage{graphicx}
\usepackage{amsmath}
\usepackage{amssymb}

\usepackage{tabu}
\usepackage{threeparttable}
\usepackage{multirow}
\usepackage{booktabs}
\usepackage{amsopn}

\usepackage{subfigure}
\usepackage{epsfig}
\usepackage{graphicx}

\usepackage{amssymb}

\usepackage{bbding}
\usepackage{pifont}
\usepackage{wasysym}

\usepackage[pagebackref=true,breaklinks=true,colorlinks,bookmarks=false]{hyperref}

\def\cvprPaperID{2869} 
\def\confYear{CVPR 2021}
\begin{document}

\title{Iterative Shrinking for Referring Expression Grounding Using Deep \\ Reinforcement Learning}

\author{Mingjie Sun\textsuperscript{1,2},~~
	Jimin Xiao\textsuperscript{1,}\thanks{corresponding author},~~
	Eng Gee Lim\textsuperscript{1}\\
{\textsuperscript{1}Xi'an Jiaotong-Liverpool University,~~}
{\textsuperscript{2}University of Liverpool~~} \\
{\tt\small mingjie.sun@liverpool.ac.uk, \{jimin.xiao,enggee.lim\}@xjtlu.edu.cn}
}

\maketitle
\pagestyle{empty}
\thispagestyle{empty}

\begin{abstract}
\footnotetext[1]{The work was supported by National Natural Science Foundation of China under 61972323 and Key Program Special Fund in XJTLU under KSF-T-02, KSF-P-02.\\\url{https://github.com/insomnia94/ISREG}}
In this paper, we are tackling the proposal-free referring expression grounding task, aiming at localizing the target object according to a query sentence, without relying on off-the-shelf object proposals. Existing proposal-free methods employ a query-image matching branch to select the highest-score point in the image feature map as the target box center, with its width and height predicted by another branch. Such methods, however, fail to utilize the contextual relation between the target and reference objects, and lack interpretability on its reasoning procedure. To solve these problems, we propose an iterative shrinking mechanism to localize the target, where the shrinking direction is decided by a reinforcement learning agent, with all contents within the current image patch comprehensively considered. Besides, the sequential shrinking processes enable to demonstrate the reasoning about how to iteratively find the target. Experiments show that the proposed method boosts the accuracy by 4.32\% against the previous state-of-the-art (SOTA) method on the RefCOCOg dataset, where query sentences are long and complex with many targets referred by other reference objects.
\end{abstract}

\section{Introduction}
\begin{figure}[t]
	\begin{center}
		\includegraphics[width=1\linewidth]{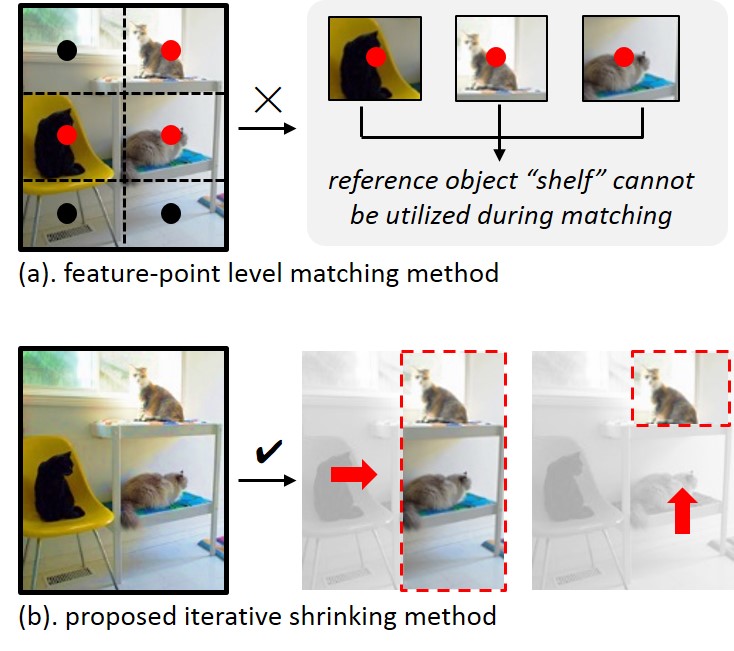}
	\end{center}
	\caption{Illustration of the difference between (a) existing feature-point level matching methods, and (b) the proposed iterative shrinking method. The query sentence for the target in this image is ``the cat above the shelf''. In (a), dots depict the feature points within the image feature map, with the surrounding area representing its corresponding image region. Red dots indicate the feature points whose corresponding image regions may cover the target (i.e., the cat on the shelf) or distracting objects (i.e., cats on other places), while black dots are unrelated feature points. It can be observed that there is no such feature point whose image region can cover both the target object (i.e., cat) and the reference object (i.e., shelf). In (b), for each iteration, the area in the dashed box indicates the image patch after shrinking, and the red arrow depicts the shrinking direction, which is decided by comprehensively considering all contents in the image patch before shrinking.}
	\label{intuitive}
\end{figure}

The aim of referring expression grounding (REG) is to recognize and localize the target object in an image according to its query sentence (referring expression), which requires joint comprehension in both visual and linguistic domains. REG is a fundamental multi-modality task, serving as the basis for many downstream tasks, including the visual question answering \cite{Jiang_2020_CVPR,Chen_2020_CVPR,kim2020modality,wang2020general,le2020hierarchical,bansal2020visual,tang2020semantic}, image caption \cite{guo2020normalized,chen2020say,cornia2020meshed,pan2020x,zhou2020more,tran2020transform,sammani2020show}, image-text matching \cite{liu2020graph,zhou2020more,chen2020imram,zhang2020context,chen2020adaptive,wang2020consensus,li2019visual}, etc. It attracts much attention from both computer vision (CV) and natural language processing (NLP) areas in recent years.

Conventional methods formulate REG as a region-retrieval problem \cite{yu2018mattnet,yu2017joint,yang2019cross,wang2019neighbourhood,liu2019learning,liu2019adaptive,gupta2020contrastive}, with proposals of all candidate objects provided in advance. The candidate proposals can be either provided from the bounding box ground-truth or generated by a pre-trained object detector (e.g., Faster RCNN \cite{ren2015faster}). The matching networks of these REG models predict the similarity score between the query and each proposal, with the highest-score proposal selected as the target object. In this way, conventional REG methods highly rely on the ground-truth bounding boxes or an accurate object detector which requires a lot of extra data to train it in advance. To overcome these shortcomings, some proposal-free methods attempt to predict the REG result without object proposals.

All existing proposal-free REG methods \cite{liao2020real,yang2019fast,sadhu2019zero,yang2020improving} directly follow the pipeline of one-stage detector (e.g., YOLOv3 \cite{redmon2018yolov3}) and adopt a two-branch style network, with the first branch to calculate the similarity score between the query sentence and each feature point within the image feature map, and the second branch to generate its bounding box coordinates. The predicted bounding box of the feature point with the highest score is viewed as the final result. 

These methods, with the matching process conducted between the query sentence and each image feature point, perform well for simple queries with the target described by its own attribute (e.g., ``man in blue''), but it is difficult for them to deal with complex queries, especially when the target is referred by another reference object. As can be observed from Fig.\ref{intuitive}.(a), since there is no such a feature point whose corresponding image region covers both the target ``cat'' and the reference object ``shelf'', the contextual relation between the target and the reference object, serving as the key to distinguish the target from other distracting objects, cannot be fully utilized in such a matching process. Thus, the REG performance is greatly compromised. 

Another shortcoming of these methods is lack of interpretability. As the matching network predicts the final matching scores of all feature points in one step, its inner reasoning procedure is hidden and invisible. Thus, if the matching model fails, it is formidable to analyze the causes. 

To tackle these problems simultaneously, we formalize REG as a sequence of image-level shrinking processes. Within each iteration, the image shrinks along a certain direction, with a non-target image region removed. After several shrinking iterations, only the target image region remains, and it is viewed as the final result. The shrinking direction in each iteration is predicted by a trainable network. As the optimal shrinking direction is uncertain for each iteration, conventional supervised training is not suited. We propose to model it as a Markov decision process \cite{jaakkola1995reinforcement} and adopt reinforcement learning (RL) to tackle it. RL only requires a ``feedback'' or ``reward'' after each shrinking step, rather than the exact supervision label. Besides, RL considers not only the current reward but also the potential reward in future, which further improves the performance.

Our proposed method can better utilize the contextual relation between the target and reference objects because the shrinking direction is decided by comprehensively considering all objects within the current image patch.
For an instance, as shown in Fig.\ref{intuitive}.(b), in the first iteration, the image region where a black cat lies on the chair is removed (i.e., shrinking towards right), because ``chair'' does not occur in the query. In the second iteration, after analyzing the spatial relation between each candidate cat and the reference object ``shelf'', it is decided to remove the image region where a cat lies inside the shelf (i.e., shrinking towards top), due to its mismatching spatial relation against the query. Ultimately, only the image region accurately covering the target cat remains. As the relation information in a query is fully utilized, the proposed method can better deal with complex queries and images (e.g., RefCOCOg \cite{mao2016generation}), whilst existing matching methods fail to handle such cases (Fig.\ref{intuitive}.(a)).

To sum up, this paper has three main contributions:

\begin{itemize}
	\item
	We make the earliest attempt to solve REG as a conditional
	decision-making process and build the first RL-based REG framework. Thanks to these iterative decisions, the reasoning about how to localize the target can be visualized step by step.    
	\item
	We formalize REG as a sequence of image-level shrinking processes, with the shrinking direction in each iteration decided by an RL agent after comprehensively considering all objects in the image patch, allowing the relation information in a query to be fully utilized.
	\item
	The proposed method boosts the accuracy by 4.32\% against the previous SOTA method \cite{liao2020real} on the RefCOCOg dataset, where query sentences are long and complex, with many targets referred by other reference objects. An average accuracy gain of 1.33\% on RefCOCOg, RefCOCO+ and RefCOCO is also achieved. 
\end{itemize}

\begin{figure*}[ht]
	\centering
	\includegraphics[width=1 \linewidth]{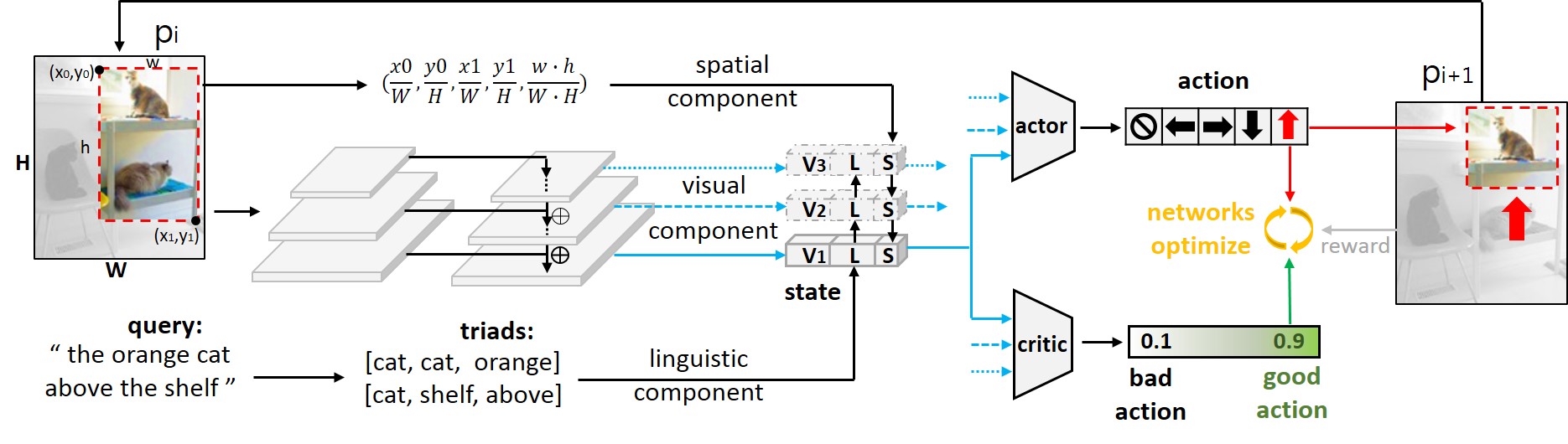}
	\caption{Architecture of our method where the blue lines indicate the states corresponding to different scale levels, and the yellow circle represents the optimizing of the actor and critic networks within each iteration.}
	\label{framework}
\end{figure*}

\section{Related Work}

\subsection{Referring Expression Grounding}
REG can be classified into two categories, including proposal-based and proposal-free. For many early proposal-based REG methods, the entire query sentence is directly encoded through a single language embedding network (e.g., LSTM \cite{hochreiter1997long}), without considering the variance among different types of query sentences \cite{rohrbach2016grounding, wang2016learning,mao2016generation,yu2016modeling}. To solve this problem, Yu \textit{et al.} \cite{yu2018mattnet} propose an attention mechanism to decompose the query sentence into three linguistic components, related to subject appearance, location and relationship to other objects. The ultimate matching result is generated after comprehensively considering all three components. To prevent the attention mechanism from only focusing on the most dominant features of both modalities, an erasing mechanism is proposed in \cite{liu2019improving}, where the most dominant linguistic or visual information is discarded to drive the model to discover more complementary linguistic-visual correspondences. 

In the proposal-free REG setting, Sadhu \textit{et al.} \cite{yang2019fast} combine the linguistic feature of the query sentence and the visual feature from the image to generate a multi-modal feature map, and predict the binary classification scores (foreground or background) for all points and their corresponding bounding box coordinates. RCCF \cite{liao2020real} adopts a correlation filter to calculate the similarity between the query sentence and each point in the image feature map, and the point with the highest similarity score is viewed as the center of the target object. Then, the size and offset are predicted by another network to generate the final bounding box for the target.

The aforementioned proposal-free REG methods \cite{yang2019fast, liao2020real} only consider a local image region and ignore the contextual object relations. Differently, we propose an iterative shrinking mechanism to predict the target object location, where the shrinking direction within each iteration is decided by comprehensively considering all objects in the current image, allowing the relation information in a query to be better utilized.

\subsection{Deep Reinforcement Learning}
Deep reinforcement learning algorithms have achieved great success on sequential decision-making problems. ``Actor-critic''~\cite{konda2000actor} is a popular RL framework that comes from several previous RL frameworks including deep Q-learning~\cite{mnih2013playing} and policy gradient~\cite{sutton2000policy} which are valued-based and policy-based strategies, respectively. 

RL has been applied to many multi-modal tasks. Rennie \textit{et al.} \cite{rennie2017self} apply the policy gradient to train the image caption network based on its baseline method \cite{xu2015show}. Specifically, the image caption network is viewed as an RL agent, whose action is to predict the next word. The state of the RL agent is the cells and hidden variables of the LSTM, as well as their corresponding attention weights. The reward function is designed according to the sentence-level evaluation metric, CIDEr score \cite{vedantam2015cider}, representing the difference between the generated sentence and the ground-truth one. Yu \textit{et al.} \cite{yu2017joint} adopt RL to guide the referring sentence generator towards less ambiguous expression. Different from \cite{rennie2017self}, the reward function is designed according to the similarity score between the generated sentence and the object image patch, calculated by a pre-trained region-language matching network. 

However, to the best of our knowledge, RL has never been used to tackle the REG task. All existing REG methods follow the static region-sentence matching strategy, leaving no space to make any sequential decisions. In this paper, we abandon the traditional matching strategy, and make the first attempt to formulate REG as a sequential shrinking processes, where an RL agent is trained to predict the shrinking direction within each iteration.

\section{Methodology}
\subsection{Overview}
Given the image $img$ and the query sentence $q$, the target object is localized through a sequence of shrinking iterations, with $i$ indicating the index of each iteration. Specifically, for a certain iteration $i$, the current image patch $p_i$ shrinks along a certain direction, decided by the RL agent, into a smaller patch $p_{i+1}$, with a non-target region eliminated. After several sequential iterations, all non-target regions are removed and only the region exactly covering the target object remains, which is viewed as the final REG result. Note that $p_1=img$ at the beginning. 

In terms of the shrinking direction selection, we adopt an RL agent to make the decision within each iteration, and view the aforementioned shrinking procedure as a Markov decision process (MDP) \cite{jaakkola1995reinforcement}. Specifically, within a certain iteration $i$, the state $s_i$ is generated according to the information embedded in $p_i$ and $q$ (Section \ref{AgentState}). Then, according to $s_i$, the RL agent predicts an action $a_i$ (shrinking direction or stop). After $a_i$ is executed, a reward $r_i$ is calculated as a ``feedback'' of $a_i$ (Section \ref{ActionReward}). Ultimately, the RL agent is trained through an actor-critic RL framework, where the RL agent consists of two components, including an actor network to predict the action, and a critic network to evaluate the predicted action (Section \ref{ActorCritic}). The architecture of our proposed method is shown in Fig.\ref{framework}.

\subsection{Agent State}
\label{AgentState}

In conventional MDP, states indicate the observations that an agent receives from the external environment. Specifically to our method, for a certain image patch $p_i$ within iteration $i$, the state $s_i$ is the input passed into the RL agent to predict the action, consisting of three components, including the linguistic feature $f^{l}_{q}$ extracted from $q$, visual feature $f^{v}_{p_i}$ extracted from $p_i$, and spatial feature $f^s_{p_i}$ extracted from the relative position of $p_i$ within $img$.

\subsubsection{Linguistic Component}
To extract the linguistic feature $f^{l}_{q}$, $q$ is first parsed into multiple discriminative triads $\{t_k\}_{k=1}^{M}$ \cite{9353247}. Each triad represents a piece of discriminative information to distinguish the target from the distracting or reference objects. As can be observed from Table \ref{triad}, a discriminative triad $t_k$ consists of three components, including a target unit $t^t_k$, a reference unit $t^r_k$ and a discriminative unit $t^d_k$, denoting the target object, the reference object against the target object and the discriminative relation between them, respectively. Note that, the discriminative unit $t^d_k$ is defined in a broad sense to take full advantage of the potential discriminative information hidden in queries. As for an instance, when it comes to ``orange cat'' in Table \ref{triad}, this query implies that there are cats in other colors with a high probability. Thus, ``cat'' is set as its reference unit to represent the distracting cats in other colors.  

For its implementation, an off-the-shelf NLP processing toolbox (e.g., Stanford CoreNLP \cite{chen2014fast} or Spacy \cite{honnibal2017spacy}) is employed to analyze the tree structure of each query, as well as the dependency label of each word in the query. In general, triads are generated within two steps. The first step is to extract the rightmost Normal Noun (NN) of the bottom-left Noun Phrase (NP) in the sentence tree as the target unit shared by all triads. The second step is to generate the reference and discriminative units using different parsing patterns. As for an instance, the subject-relation-object phrase, such as ``the man holding a cat'', firstly the second component of a particular dependency, whose first component is the same as the target unit, and the dependency type is nsubj, is viewed as a discriminative unit. Then, the second component of another dependency, whose first component is the same as that of the current discriminative unit, is viewed as its corresponding reference unit.    

Then, a pre-trained word vector generation method (e.g., Word2vec \cite{mikolov2013efficient} or Glove \cite{pennington2014glove}) is adopted to extract the linguistic embedding $f^{l}_{t^t_k}$, $f^{l}_{t^r_k}$, $f^{l}_{t^d_k}$, corresponding to $t^t_k$, $t^r_k$, $t^d_k$, respectively. The linguistic feature $f^{l}_{t_k}$ of a certain triad $t_k$ is the concatenation of its three components' linguistic features:
\begin{equation}
f^{l}_{t_k} = f^{l}_{t^t_k} \oplus f^{l}_{t^r_k} \oplus f^{l}_{t^d_k},
\end{equation}
where $\oplus$ indicates the concatenation operation. Similarly, another concatenation operation is further conducted on all triads' linguistic features:
\begin{equation}
f^{l}_{q} = f^{l}_{t_1} \oplus f^{l}_{t_2} \oplus \cdots \oplus f^{l}_{t_M},
\end{equation}
and $f^{l}_{q}$ represents the ultimate linguistic feature of $q$. Note that, if the total number of the triads parsed from a query is less than $M$, the first triad feature $f^{l}_{t_1}$ is repeated until the triad number reaches $M$. 

\begin{table}[]
	\caption{Common queries and their corresponding discriminative triads on the RefCOCOg, RefCOCO+ and RefCOCO datasets. \textbf{ID} is the index of a discriminative triad for a query, as a complex query may generate multiple discriminative triads. \textbf{T/R/D} indicates the target, reference, and discriminative unit, respectively.}
	\begin{center}
		\begin{tabular}{|p{95pt}<{\centering} |p{7pt}<{\centering} |p{25pt}<{\centering} p{25pt}<{\centering} p{30pt}<{\centering}|}
			\hline
			\textbf{Query} & \textbf{ID} & \textbf{T} & \textbf{R}  & \textbf{D}\\
			\hline
			``lady''         & 1   & lady     & lady    & SELF  \\
			\hline
			``left''              & 1   & UKN     & UKN    & left  \\
			\hline
			``left lady''       & 1  & lady     & lady    & left  \\
			\hline
			``orange cat''     & 1   & cat     & cat    & orange  \\
			\hline
			``cat above a shelf''    & 1  & cat     & shelf    & above  \\
			\hline
			``lady holding a cat''     & 1 &  lady    & cat    & holding  \\
			\hline
			\multirow{5}*{\shortstack{``the left lady in white \\ holding an orange cat \\and standing on a table''}}     & 1  & lady     & lady    & left  \\
			~       & 2  & lady     & lady    & white   \\
			~       & 3  & lady     & table    & on     \\
			~       & 4  & lady     & cat    & holding  \\
			~       & 5  & cat     & cat    & orange    \\
			\hline
		\end{tabular}
	\end{center}
	\label{triad}
\end{table}

\subsubsection{Visual and Spatial Components}

The visual feature of the image patch $p_i$ is extracted through an off-the-shell convolutional network (e.g., VGG \cite{simonyan2014very} or ResNet \cite{he2016deep}) and the feature pyramid network (FPN) \cite{lin2017featur}. Specifically, the convolutional network, e.g., ResNet \cite{he2016deep}, first generates three raw visual feature maps, $f^{\tilde{v_1}}_{p_i}$, $f^{\tilde{v_2}}_{p_i}$, $f^{\tilde{v_3}}_{p_i}$, with different scales of $[28 \times 28 \times 512]$, $[14 \times 14 \times 1024]$ and $[7 \times 7 \times 2048]$, respectively. Then, the up-sampling, fusing and average-pooling operations of FPN \cite{lin2017featur} are conducted on these three raw feature maps, which generates three one-hot vectors with the same dimension ($D_v$), $f^{v_1}_{p_i}$, $f^{v_2}_{p_i}$ and $f^{v_2}_{p_i}$, as the corresponding visual feature for each scale level.

The spatial feature $f^s_{p_i}$ of the image patch $p_i$ is encoded by a 5-D vector, $[\frac{x_{tl}}{W}, \frac{y_{tl}}{H}, \frac{x_{br}}{W}, \frac{y_{br}}{H}, \frac{w\cdot h}{W\cdot H}]$, where $[x_{tl}, y_{tl}]$ and $[x_{br}, y_{br}]$ refer to the top-left and bottom-right coordinates of $p_i$ within $img$, respectively. $w$ and $h$ indicate the width and height of $p_i$, while $W$ and $H$ denote the width and height of $img$. $f^s_{p_i}$ can provide the relative position of $p_i$ within $img$, through which the RL agent can infer the relative location of the target object within $img$.

The ultimate state $s_i$ consists of three sub-states $s^c_i$ with $c\in \{1,2,3\}$, each of which corresponds to a certain scale level $c$. Each sub-state is designed as the concatenation of the corresponding linguistic, visual and spatial features:
\begin{equation}
s^c_i = f^{l}_{q} \oplus f^{v_c}_{p_i} \oplus f^{s}_{p_i}.
\end{equation}

\begin{figure}[t]
	\begin{center}
		\includegraphics[width=1 \linewidth]{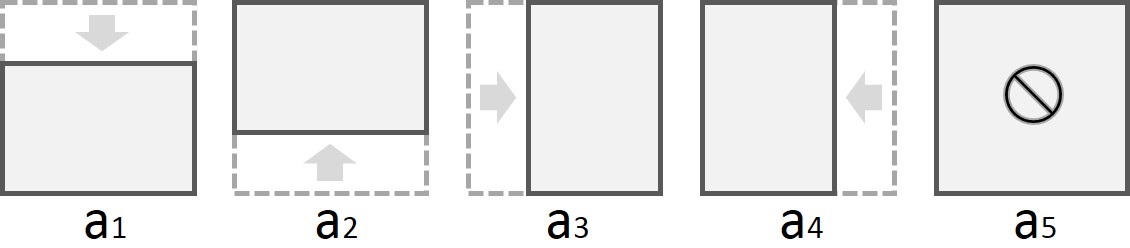}
	\end{center}
	\caption{Illustration of our action design. $\{a^1,a^2,a^3,a^4\}$ corresponds to image shrinking toward 4 different directions, respectively. $a^5$ corresponds to the STOP action.}
	\label{actions}
\end{figure}

\subsection{Action and Reward}
\label{ActionReward}

In MDP, after observing the environment, an agent takes actions to interact with the external environment. A reward is given to provide the feedback of the conducted action. As shown in Fig.\ref{actions}, within a certain iteration $i$, the RL agent selects an action $a_i$ among the action pool $A$, consisting of 5 candidate actions, $\{a^1,a^2,a^3,a^4,a^5\}$. The first 4 candidate actions, $\{a^1,a^2,a^3,a^4\}$, indicate the top side of $p_i$ shrinks towards down, the bottom side of $p_i$ shrinks towards up, the left side of $p_i$ shrinks towards right and the right side of $p_i$ shrinks towards left, respectively. $a_5$ indicates the end of the shrinking process (i.e., $p_i=p_{i+1}$), and the current image patch is viewed as the final REG result.  

To naturally introduce the shrinking stride decay in an episode, the value of the shrinking stride is decided according to the width and height of the image patch $p_i$ before shrinking:
\begin{equation}
\begin{split}
\hat{y}_{tl} = y_{tl} + \alpha (y_{br} - y_{tl}), \ \mathrm{if} \ a_i=a^1 \\
\hat{y}_{br} = y_{br} - \alpha (y_{br} - y_{tl}), \ \mathrm{if} \ a_i=a^2 \\
\hat{x}_{tl} = x_{tl} + \alpha (x_{br} - x_{tl}), \ \mathrm{if} \ a_i=a^3 \\
\hat{x}_{br} = x_{br} - \alpha (x_{br} - x_{tl}), \ \mathrm{if} \ a_i=a^4,
\end{split}
\end{equation}    
where $[x_{tl},y_{tl}]$ and $[x_{br},y_{br}]$ indicate the coordinates of the top-left and bottom-right points of $p_i$ within $img$. $[\hat{x}_{tl},\hat{y}_{tl}]$ and $[\hat{x}_{br},\hat{y}_{br}]$ indicate the coordinates of the top-left and bottom-right points of $p_{i+1}$ within in $img$. In this way, at the early stage of an episode, relatively large shrinking strides are adopted to quickly ``move'' towards to the target. While in the late stage, relatively small shrinking strides are adopted to meticulously localize the accurate position of the target.

To predict such an action $a_i$, the state $s^c_i$ for each scale level is passed into the RL agent respectively, generating three action probability vectors, with the average one serving as the final probability vector. The action with the highest probability is selected as $a_i$. 

The reward $r_i$ is designed according to the overlap between the image patch after shrinking and the ground-truth bounding box, as follows:
\begin{equation} r_i=\left\{\begin{matrix}
0 & IoU < 0.3 \ \ or \ \  \Delta_{IoU} \leqslant 0 \\ 
1 & 0.3 \leqslant IoU < 0.5 \ \ and \ \  0 < \Delta_{IoU} \\
10 & 0.5 \leqslant IoU \ \ and \ \  0 < \Delta_{IoU}
\end{matrix}\right.
\label{reward},
\end{equation}
where $IoU$ refers to the intersection-over-union between the image region after shrinking and the ground-truth bounding box of the target, and $\Delta_{IoU}$ indicates the $IoU$ difference between the current iteration and the previous iteration. We set major difference between the good and bad action's rewards, so as to facilitate the training of the RL agent.

\begin{table*}[]
	\caption{Accuracy comparison against other REG methods. ``PF'' indicates proposal-free REG methods. Specifically, methods on the top half of the table are not proposal-free, while methods on the bottom half are proposal-free. ``Average'' is the average accuracy of RefCOCOg,  RefCOCO+, and RefCOCO, where mean accuracy of testA and testB is used for RefCOCO+ and RefCOCO. The method with the highest accuracy is in bold.}
	\begin{center}
		\begin{tabular}{|p{90pt}|p{20pt}<{\centering}|c|p{30pt}<{\centering}|c|p{30pt}<{\centering}|p{30pt}<{\centering}|p{30pt}<{\centering}|p{30pt}<{\centering}|}
			\hline
			\multirow{2}{*}{Method}            & \multirow{2}{*}{PF}          & \multirow{2}{*}{Visual Extractor} & \multirow{2}{*}{Average} & RefCOCOg & \multicolumn{2}{c|}{RefCOCO+} & \multicolumn{2}{c|}{RefCOCO} \\ \cline{5-9} 
			&                                   &                                   &                       & test     & testA         & testB         & testA         & testB        \\ \hline \hline
			SLR\cite{yu2017joint} \tiny{CVPR17} &                                   & ResNet-101                         & 61.25                 & 59.63    & 60.74         & 48.80         & 73.71         & 64.96        \\
			DGA\cite{yang2019dynamic} \tiny{ICCV19}                         &                                   & VGG-16                             & 65.26                 & 63.28    & 69.07         & 51.99         & 78.42         & 65.53        \\
			MAttNet\cite{yu2018mattnet} \tiny{CVPR18}                     &                                   & ResNet-101                         & 68.33                 & 67.01    & 70.26         & 56.00         & 80.43         & 69.28        \\ \hline
			FAOAVG\cite{yang2019fast} \tiny{ICCV19}                      & \multirow{4}{*}{\ding{52}}    & DarkNet-53                         & 61.79                 & 57.45    & 60.56         & 52.86         & 74.81         & 67.59        \\
			MCN\cite{luo2020multi} \tiny{CVPR20}                         &                                   & VGG-16                             & 65.69                 & 62.29    & 65.24         & 54.26         & 76.97         & \textbf{73.09}        \\
			RCCF\cite{liao2020real} \tiny{CVPR20}                        &                                   & DLA-34                             & 67.30                 & 65.73    & 70.35         & 56.32         & \textbf{78.93}         & 66.73        \\
			Ours                               &                                   & ResNet-101                         & \textbf{68.63}                 & \textbf{70.05}    & \textbf{71.05}         & \textbf{58.25}         & 74.27         & 68.10       \\ \hline
		\end{tabular}
		\begin{tablenotes} 
		\item * RefCOCO results of RCCF are from Table 4, setting 7 in \cite{liao2020real}, where the feature backbone is obtained without the detection pretraining using the additional classification labels of the 80 categories in MS-COCO dataset. As the results of RefCOCOg and RefCOCO+ are not provided in Table 4, results from Table 3 in \cite{liao2020real} are used instead, but the comparison is actually unfavorable for the proposed method. The darknet-based MCN \cite{luo2020multi} is not included in this table for the same reason.  
	\end{tablenotes} 
		\label{comparisonCOCO}
	\end{center}
\end{table*}

\subsection{Actor-critic Training}
\label{ActorCritic}
The proposed RL agent is designed and optimized under the ``actor-critic'' framework \cite{konda2000actor}, consisting of two networks, including the actor network to predict the action among the candidate action pool, and the critic network to predict the score of a certain state, which is further utilized to evaluate the quality of the predicted action. During each iteration of training, the actor and critic networks are optimized in order, while in the inference stage, only the actor network is used to predict the action to localize the target.

Specifically, the actor network is first optimized in a policy-based way:
\begin{equation} 
\theta  = \theta{}' + \l_a   \nabla (log\pi_{\theta{}'} (s_{i},a_{i}))   A(s_{i},a_{i}), 
\label{actor}
\end{equation}
where
\begin{equation}
A(s_{i},a_{i}) = \delta_i =  r_i + \gamma V_{w{}'}(s_{i+1})-V_{w{}'}(s_i). 
\label{advantage}
\end{equation}
In Eq.(\ref{actor}) and Eq.(\ref{advantage}), $\theta{}'$ and $\theta$ indicate the weights of the actor network before and after the update. $l_a$ is the learning rate corresponding to the actor network. $\pi(s,a)$ is the actor network's policy function to predict the probability of taking action $a$ for state $s$. $A(s,a)$ is the advantage function, which is equal to the TD error function in the ``actor-critic'' framework \cite{mnih2016asynchronous}. $r_i$ is the reward calculated using Eq.(\ref{reward}), and $\gamma$ refers to the discount factor. $w{}'$ is the weight of the critic network before the update of the critic network. $V(s)$ indicates the predicted score for state $s$, generated by the critic network. 

After the actor network is updated, the critic network is optimized in a value-based way:
\begin{equation}
w = w{}' + \l_c \delta_i \nabla_{w{}'} V_{w{}'}(s_i). 
\label{criticupdate}
\end{equation}
In Eq.(\ref{criticupdate}), $w{}'$ and $w$ indicate the weights of the critic network before and after the optimization.  $l_c$ is the learning rate corresponding to the critic network. $\delta_i$ is the TD error function calculated in Eq.(\ref{advantage}).  

In this way, the joint optimization of the actor and critic networks can avoid the disadvantages of both value-based and policy-based methods during training. In other words, the proposed RL agent can be optimized after each iteration, rather than waiting until the end of each episode, which dramatically speeds up the training process yet maintains training stability.

\subsection{Implementation Details}
\label{implementation}
To generate the state, Stanford CoreNLP \cite{chen2014fast} is adopted to parse the query sentence into triads, and Glove word vector \cite{pennington2014glove} is adopted to generate the linguistic feature of each triad. The triad number $M$ for each query is set as 2, because a too small $M$ cannot contain enough information from queries, while an extremely large $M$ may greatly increase the network complexity. Resnet101 \cite{he2016deep} is used to generate the raw visual feature of the image patch, and FPN \cite{lin2017featur} is adopted to generate the fused visual feature, with $D_v$ set as 512. In terms of the action, $\alpha$ is set as 0.2 to generate a moderate shrinking stride. During the training, $\gamma$ is set as 0.9, and the learning rate is set as 1e-7 for both the actor and critic networks. ADAM \cite{kingma2014adam} is adopted as the optimizer. Training is conducted on one Titan RTX GPU for around 800,000 episodes.

Similar as \cite{liao2020real}, to obtain a more precise bounding box for a target, we employ a post-processing box refinement model after the shrinking process, where the input is $img$ and the predicted bounding box from the shrinking process, and the output is the new refined bounding box. Such box refinement network is trained using the L2 loss between the coordinates of a ground-truth box in MS-COCO \cite{lin2014microsoft} and the coordinates of corresponding sampled box obtained by adding Gaussian noise to it.

\begin{table}[]
	\caption{Performance comparison of our method against other REG methods on the RefCLEF dataset. The method with the highest accuracy is in bold.}
	\begin{center}
		
		\begin{tabular}{|p{65pt}|p{60pt}<{\centering}|}
			\hline
			Method  & Accuracy \\ \hline \hline
			CITE\cite{plummer2018conditional}\tiny{ECCV18}       & 34.13    \\
			IGOP\cite{yeh2017interpretable}\tiny{NIPS17}       & 34.70    \\
			RCCF\cite{liao2020real}\tiny{CVPR20}     & 63.79    \\ \hline
			Ours                   & \textbf{65.48}    \\ \hline
		\end{tabular}
		\label{comparisonCLEF}
	\end{center}
\end{table}

\section{Experiments}
\subsection{Datasets and Metrics}
The proposed method is evaluated on four common datasets including RefCOCO \cite{yu2016modeling}, RefCOCO+ \cite{yu2016modeling}, RefCOCOg \cite{mao2016generation} and RefCLEF \cite{kazemzadeh2014referitgame}. The first three datasets are collected from MS-COCO \cite{lin2014microsoft}. The main difference between RefCOCO+ and RefCOCO is that the former one forbids the absolute location words when generating the query sentences, requiring more attention to the appearance discrimination. RefCOCOg contains longer expressions for both appearance and locations. Referring sentences in RefCLEF are annotated casually, so it is normally used as an auxiliary dataset. The metric used to evaluate the accuracy of the proposed method is similar to the object detection task, where the intersection-over-union (IoU) between the predicted bounding box and the ground-truth one is calculated. Then, the one with IoU higher than 0.5 is treated as correct.

\subsection{Comparison with State-of-the-Art}
To evaluate our proposed method, we compare it with other SOTA REG methods, which are divided into two groups. The first group, shown in the top half of Table \ref{comparisonCOCO}, consists of REG methods with object proposals, including SLR \cite{yu2017joint}, DGA \cite{yang2019dynamic} and MAttNet \cite{yu2018mattnet}. The second group, shown in the bottom half of Table \ref{comparisonCOCO}, is composed of proposal-free REG methods, including FAOAVG \cite{yang2019fast}, MCN \cite{luo2020multi} and RCCF \cite{liao2020real}.
\begin{table}[]
	\caption{Ablation studies on RefCOCO (testA) and RefCOCOg (test) datasets. The method with the highest accuracy is in bold.}
	\begin{center}
		\begin{tabular}{|p{90pt}<{\centering}|p{40pt}<{\centering}|p{45pt}<{\centering}|}
			\hline
			Setting                & RefCOCO  & RefCOCOg        \\ \hline \hline 
			w/o triad              & 67.52      & 60.17         \\
			w/o multi-scale        & 73.33      & 66.35         \\
			w/o spatial feature    & 62.42      & 54.88         \\ \hline 
			fixed stride           & 45.70      & 39.46         \\
			w/o refinement             & 70.18      & 65.62         \\ \hline
			supervised training    & 63.84      & 61.36         \\ \hline
			ours                   & \textbf{74.27}      & \textbf{70.05}    \\
			\hline
			
		\end{tabular}
		\label{ablation}
	\end{center}
\end{table}
As can be observed from Table \ref{comparisonCOCO}, the proposed method achieves a large accuracy gain (4.32\%) over the previous SOTA method, RCCF \cite{liao2020real}, on the test set of the RefCOCOg dataset. When the backbone is replaced by VGG, the proposed method achieves 66.32\% on this dataset. The cause of such big accuracy gain is that, queries in RefCOCOg are much longer and more complex than queries in other datasets, with many targets referred by other neighboring reference objects, namely ``targets with reference''. Thus, the key to localize such targets is to take full advantage of the relation information between the target and reference objects in queries, which is difficult for the feature-point level matching mechanism \cite{yang2019fast,luo2020multi,liao2020real}. Nevertheless, the image-level shrinking mechanism allows the proposed method to consider all objects within the image patch, and better utilize such contextual relation information, so as to achieve higher accuracy.
\begin{figure}
	\centering
	\includegraphics[width=0.9 \linewidth]{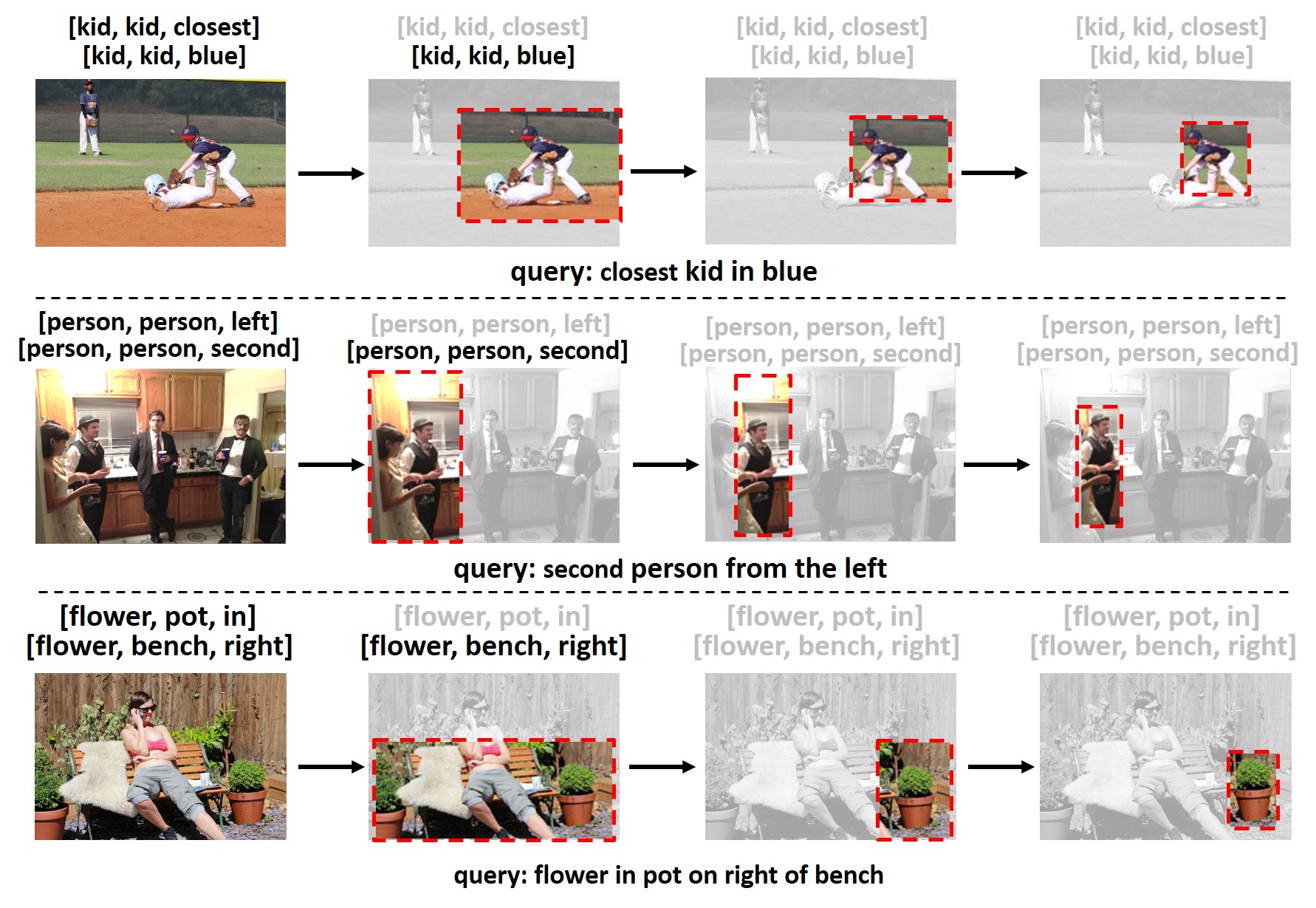}
	\caption{Visualization of the reasoning process about how to iteratively localize the target. The image patch generated through shrinking is marked in red box. The parts above the images indicate the triads parsed from the query, with the black units representing the contents remaining in the corresponding image patch, and the gray units indicating that the visual contents of the corresponding distracting or reference objects are already removed.}
	\label{visualization}
\end{figure}

\begin{figure}
	\centering
	\includegraphics[width=0.7 \linewidth]{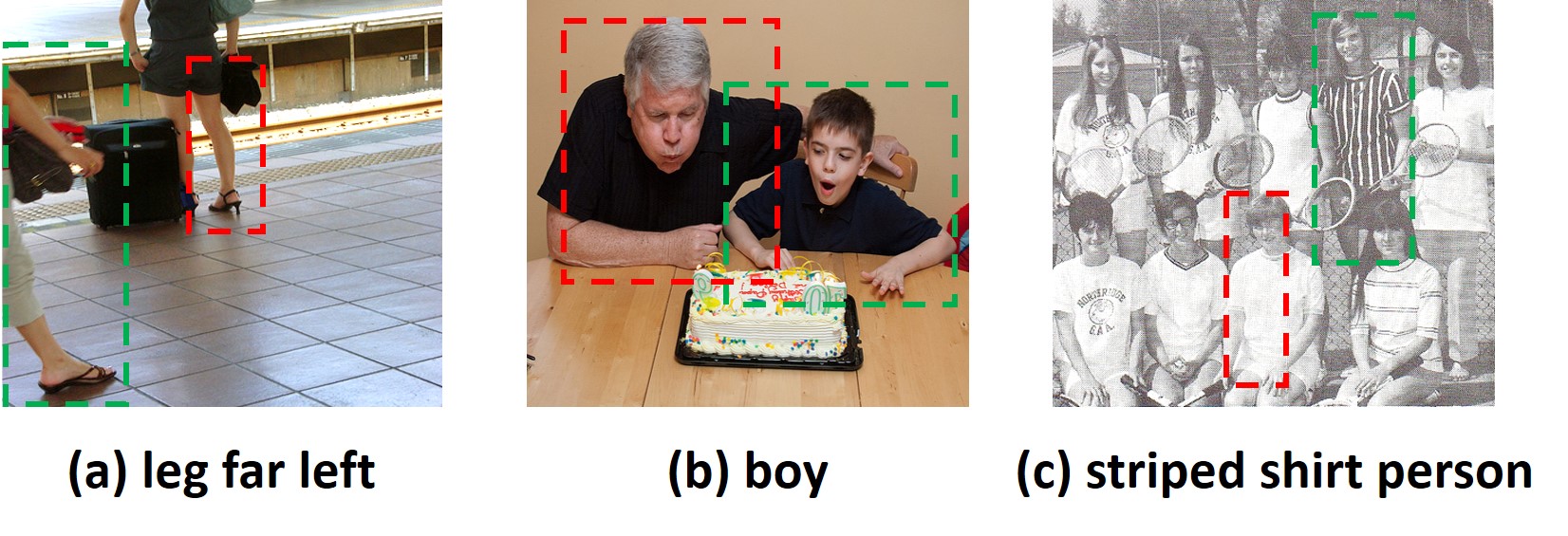}
	\caption{Some failure cases of the proposed method where the green box is the ground-truth one, and the red box is the predicted one which is incorrect.}
	\label{fail}
\end{figure}
Then, similar accuracy improvement can be seen on RefCOCO+ (1.32\%) and RefClef (1.69\%) datasets, where many queries are complicated as well, as shown in Table \ref{comparisonCOCO} and Table \ref{comparisonCLEF}. However, the accuracy gains on RefCOCO+ and RefClef are not as large as that on RefCOCOg. We believe the reason is that the proportion of ``targets with reference'' in RefCOCO+, as well as RefClef, is relatively smaller than that in RefCOCOg. Ultimately, when it comes to the RefCOCO dataset, where most queries are simple and ``targets with reference'' are rare, the proposed method only achieves a comparable accuracy, as the superiority of better utilizing the contextual relation information turns useless.

Aforementioned evaluation results on different datasets further prove the positive correlation between the accuracy gain of the proposed method and the number of ``targets with reference''. Though we have not achieved accuracy gain on all datasets, the average accuracy obtained by the proposed method, on the RefCOCOg, RefCOCO+ and RefCOCO datasets, is still higher than the previous SOTA method RCCF \cite{liao2020real} by 1.33\%.

Apart from the accuracy gain, it can also be observed from Table \ref{comparisonCOCO} that the performance of proposed method is more stable and robust on different datasets compared with other methods. For instance, RCCF \cite{liao2020real} achieves quite high accuracy (78.93\%) on testA of the RefCOCO dataset, but it does not perform well on the RefCOCOg dataset (65.73\%), with a gap of more than 13\%. The proposed method's performance is much more stable, with the corresponding accuracy variation less than 5\%, demonstrating the robustness of our proposed method.

\subsection{Visualization of Reasoning Process}
The reasoning process about how to iteratively find and localize the target can be visualized through the sequential shrinking processes, as shown in Fig.\ref{visualization}. For instance, in the first case in Fig.\ref{visualization}, the first step of the RL agent is to remove the first distracting object, the kid far from us, which is against the query information ``closest'', making its corresponding triad [kid,kid,closest] gray. Then, the second step is to eliminate the second distracting object, the kid in white, which is against the query information ``blue'', making its corresponding triad [kid,kid,blue] gray. Ultimately, after a few small-stride shrinks, the precise target region can be obtained. Note that in the third case, the triad [flower,pot,in] turns gray because the distracting flowers, which are not in the pot, are removed from the image. In general, the distracting and reference objects are iteratively excluded in the early stage, and then the target region is gradually refined, until the exact bounding box of the target is obtained.

Some failure cases of the proposed methods are also reported in Fig.\ref{fail}. In Fig.\ref{fail}.(a), the heavily-occluded target is difficult to be localized. Besides, tiny facial difference is also hard to be recognized, like the man and the boy wearing the same clothes in Fig.\ref{fail}.(b). The similar challenge is also encountered when dealing with the difference in the textual description about the clothes styles, as shown in Fig.\ref{fail}.(c). We believe the reason for these failure cases is that the similar situations are very rare in the training set, making them quite tough to be handled for REG methods. We will try to address these hard cases in our future works.

\subsection{Ablation Study}
\subsubsection{RL State Design}
The first ablation study is to explore the contribution of the linguistic, visual and spatial components in the proposed state, respectively. As shown in Table \ref{ablation}, for the linguistic component, the triad-based feature is replaced with the feature extracted through Bert \cite{devlin2018bert} (\textbf{w/o triad}). Then, the multi-scale feature maps are not utilized, with the raw feature extracted from the CNN network directly employed as the final visual feature (\textbf{w/o multi-scale}), to investigate the contribution of the multi-scale visual feature. Lastly, the spatial component is removed from the state to check its influence (\textbf{w/o spatial}). The proposed method shows an accuracy gain against all aforementioned settings, which demonstrates the significance of each component in the proposed RL state. Note that the accuracy gain against \textbf{w/o triad} comes from better feature representation for the query sentence using the triads, as trivial words in the query (e.g., articles, conjunctions) are eliminated, and the key words are organized into a unified format, allowing a better multi-modal comprehension.

\subsubsection{RL Action Design}
The second ablation study is to study different designs for the agent action. As shown in Table \ref{ablation}, firstly, the adaptive shrinking stride is replaced with the fixed shrinking stride (\textbf{fixed stride}). Specifically, within each iteration, the shrinking stride is fixed as $\alpha H$ for $a_1$ and $a_2$, and $\alpha W$ for $a_3$ and $a_4$, where $H$ and $W$ are the width and height of the original image $img$, respectively, and $\alpha$ is set as 0.2, being the same as in the proposed adaptive shrinking mechanism. The performance drops dramatically in this setting, demonstrating an adaptive shrinking stride contributes significantly to precise localization. Apart from the actions in the shrinking process, we further disable the bounding box refinement network described in Section \ref{implementation} (\textbf{w/o refinement}), with the comparison result demonstrating the contribution of the bounding box refinement step.

\subsubsection{RL or Supervised Learning?}
The third ablation study is to explore whether the proposed method can be trained in a supervised way, and we introduce \textbf{supervised training} in Table \ref{ablation}. In this setting, we randomly sample a region, which covers the  whole target object box and  its size is larger than the target object box. 
Then, one side of the sampled region, among the top, bottom, left and right sides, with the highest spatial difference against the target region, is set as the positive label indicating the shrinking direction, if the difference is greater than a threshold. Otherwise, the stop action is set as the positive label. As can be observed in Table \ref{ablation}, the proposed RL performs much better than the supervised training. We believe that the major reason is that the adopted sampling strategy for the supervised training cannot cover all situations in reality. In addition, the RL training can consider not only the current profit but also the potential profit in the future, making a higher overall performance.

\section{Conclusion}
We have proposed an iterative shrinking mechanism to tackle the REG task, where the image iteratively shrinks into the correct target region. An RL agent is adopted to decide the shrinking direction within each iteration, where all objects in the image patch, including the target, reference and distracting objects, are comprehensively considered to predict the optimal shrinking direction. In future, we plan to deceptively eliminate the contextual information whose corresponding image region has already been removed.

{
\small
\bibliographystyle{ieee_fullname}
\bibliography{egbib}
}

\end{document}